\documentclass[conference]{IEEEtran}
\IEEEoverridecommandlockouts
\usepackage{cite}
\usepackage{amsmath,amssymb,amsfonts}
\usepackage{algorithm2e}
\usepackage{algorithmic}
\usepackage{graphicx}
\usepackage{textcomp}
\usepackage{xcolor}
\def\BibTeX{{\rm B\kern-.05em{\sc i\kern-.025em b}\kern-.08em
    T\kern-.1667em\lower.7ex\hbox{E}\kern-.125emX}}
\begin{document}

\title{Reinforcement Learning for Standards Design}

\author{\IEEEauthorblockN{Shahrukh Khan Kasi}
\IEEEauthorblockA{\textit{AI4Networks Center} \\
\textit{University of Oklahoma}\\
Tulsa, OK, USA \\
shahrukhkhan.kasi@outlook.com}
\and
\IEEEauthorblockN{Sayandev Mukherjee}
\IEEEauthorblockA{\textit{Next-Generation Systems} \\
\textit{CableLabs}\\
Santa Clara, CA, USA \\
s.mukherjee@cablelabs.com}
\and
\IEEEauthorblockN{Lin Cheng}
\IEEEauthorblockA{\textit{Next-Generation Systems} \\
\textit{CableLabs}\\
Louisville, CO, USA \\
l.cheng@cablelabs.com}
\and
\IEEEauthorblockN{Bernardo A.~Huberman}
\IEEEauthorblockA{\textit{Next-Generation Systems} \\
\textit{CableLabs}\\
Santa Clara, CA, USA \\
b.huberman@cablelabs.com}
}

\maketitle

\begin{abstract}
Communications standards are designed via committees of humans holding repeated meetings over months or even years until consensus is achieved.  This includes decisions regarding the modulation and coding schemes to be supported over an air interface.  We propose a way to ``automate" the selection of the set of modulation and coding schemes to be supported over a given air interface and thereby streamline both the standards design process and the ease of extending the standard to support new modulation schemes applicable to new higher-level applications and services.  Our scheme involves machine learning, whereby a \emph{constructor} entity submits proposals to an \emph{evaluator} entity, which returns a score for the proposal.  The constructor employs reinforcement learning to iterate on its submitted proposals until a score is achieved that was previously agreed upon by both constructor and evaluator to be indicative of satisfying the required design criteria (including performance metrics for transmissions over the interface).
\end{abstract}


\section{Introduction}
The design of the physical layer (PHY) interface for a communications standard is usually done by a standards committee, comprising delegates from many member companies, institutions, and organizations.  Consensus is achieved on the elements of the PHY design only after repeated meetings stretching over months or even years.  Moreover, the committee to decide on the specifics of the PHY layer is different from the committee to decide on the applications and services to be supported and prioritized by the communications standard. Thus, if a new application or service becomes popular and the communications standards organization wishes to support it, then the PHY layer committee may need to re-initiate their extensive and exhausting meeting schedule in order to finalize the selection of modulation and coding schemes (MCSs) that will allow for this application to be supported at the desired level of performance.

Machine learning (ML) methods offer the potential to automate many activities that consume vast amounts of time from large numbers of humans.  In this work, we investigate the use of ML to automate the selection of MCS sets for communication across a given interface.  The goal is to allow MCS selection to happen given only the desired performance specification across the PHY layer that is required by the next level (say the MAC layer, or higher) of the communication stack.

In the present work, we illustrate the applicability of ML-based standards design by showing how a selection of MCS sets to fulfill certain desired criteria can emerge automatically from Reinforcement Learning (RL) applied to a ``game'' played by two ML model entities, one called the \emph{constructor} and the other the \emph{evaluator}. The constructor proposes an MCS set to the evaluator, which evaluates the proposal on a set of criteria that is not revealed to the constructor.  The evaluator returns a \emph{score} to the constructor that describes the quality of the proposed MCS set.  This score is then treated by the constructor like the \emph{reward} in RL and used by the constructor to propose and submit a revised MCS set to the evaluator, and so on.  The game ends when the evaluator returns a score that is higher than some threshold that the evaluator and constructor have both agreed upon earlier as implying satisfactory performance of the proposed MCS set.  The MCS set that achieves this score is then adopted for the standard.

Note that the threshold score could incorporate not only the performance requirements of the next higher (say MAC) layer in the communications stack, but also other criteria like which codecs are available to use royalty-free.  Note also that although we will restrict ourselves to studying the proposed scheme on the PHY layer, it is equally applicable to the MAC layer with requirements for satisfactory performance coming down from, say, the Network layer, and so on.  Thus, it is possible to conceive that standards design could be completely automated in a top-down methodology just by imposing a set of requirements at the highest (i.e., application) layer, and percolating the appropriate requirements and constraints down one layer at a time to a pair of constructor-evaluator ML models at each layer that together design the elements of the standard that satisfy the requirements for that layer.

In Sec.~\ref{sec:background} we describe the MCS selection problem in a PHY layer defined by an air interface.  For concreteness, we assume that we can select from amongst all MCS selections allowed by the LTE-A standard.  In Sec.~\ref{sec:ml} we propose our RL-based solution employing the game between a constructor and an evaluator.  We define the format of constructor's proposal and the criterion that the evaluator uses to compute the score for a submitted proposal.  The constructor then receives the score from the evaluator for its proposal.  In Sec.~\ref{sec:rl} we describe the RL scheme whereby the constructor iterates on its proposal and generates an updated MCS set to propose to the constructor, and so on.  Finally, in Sec.~\ref{sec:results}, we show the results of simulations for a constructor-evaluator pair and discuss our conclusions in Sec.~\ref{sec:concl}.    

\section{MCS selection and SIR distribution}
\label{sec:background}
\begin{figure}[htbp]
\centerline{\includegraphics[width=\columnwidth]{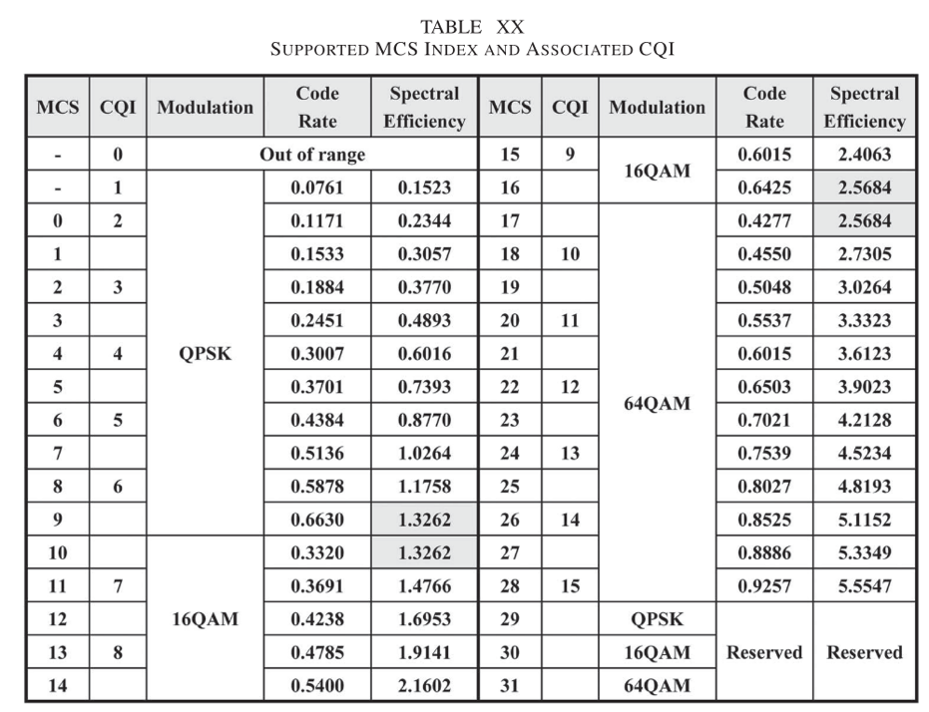}}
\caption{Lookup table for MCS, associated CQI, and SE (from~\cite{b2}).}
\label{fig:mcs_cqi}
\end{figure}

The 3GPP standard specifies a total of 29 different schemes pairing a modulation scheme with a (channel) coding rate~\cite{b1, b2} (see entries 0 through 28 in the ``MCS'' column in the table in Fig.~\ref{fig:mcs_cqi}).  Each coding rate corresponds to a spectral efficiency (SE), which in turn corresponds to a certain number of information bits transmitted per OFDM resource element (RE).  In other words, this figure of bits per RE can be used to compute the peak throughput for any arbitrary MCS (even one not in the 3GPP set of 29 MCSs) and for any arbitrary bandwidth. 

Further, extensive simulations by various 3GPP member companies~\cite{b3, b4} have provided data that can be used to create lookup tables that provide the channel quality indicator (CQI) threshold values (and corresponding SINR values over an AWGN channel) at which the MCS selection for the channel switches from one SE value to the next higher one.  In this work, we use the lookup table reported from empirical studies in~\cite{b7} for CQI to SINR mapping in our simulations.  Moreover, by interpolating between these SINR transition values for any given SE not in the table, we can estimate the SINR at the transition for any MCS not in the 3GPP set of 29 MCSs.

In other words, such lookup tables therefore let us compute the peak throughput and SE for an arbitrary MCS.  However, such tables cannot calculate or estimate the distribution of the actual (as opposed to peak) throughput experienced by an arbitrarily-located user terminal (UE in 3GPP terminology).  

For this purpose, we seek an easy-to-calculate analytical or semi-analytical expression for the SINR (or the SIR, if we make the reasonable assumption that the channel is interference-limited rather than noise-limited) that is nonetheless indicative of a real-world deployment.  In this paper, we start with the relatively simple expression for the distribution of the SIR at an arbitrarily-located UE in a deployment of (omnidirectional) base stations whose locations are the points of a homogeneous Poisson Point Process (PPP) on the plane, and the UE is served by the strongest (in terms of received power) base station (BS) at the UE. With these assumptions and assuming a path loss exponent of 4, we know that the complementary cumulative distribution function (CCDF) of the SIR at an arbitrary UE, evaluated at any argument $\gamma > 0$, is given by the approximate expression~\cite[(4.74)--(4.77)]{b5}
\begin{equation}
	(2/\pi)/\sqrt{\gamma} - (1/\pi)(1/\sqrt{\gamma}-1)^2 1_{(0,1)}(\gamma),
	\label{eq:ccdf}
\end{equation}
which is exact for $\gamma \geq 1/2$.

Note that the CCDF in~\eqref{eq:ccdf} does not depend on the density of the assumed homogeneous PPP of the BSs, nor does it depend on the maximum transmit power of the BSs.  Moreover, the CCDF~\eqref{eq:ccdf} is quadratic in $\sqrt{\gamma}$, which yields an analytic formula for the inverse of the corresponding CDF, which can therefore be used to simulate draws of SIR from this distribution for different UEs.  The advantage of this ease of simulation will be seen in the ML-based approach to MCS selection that we propose and describe next.

\section{Machine Learning for MCS selection}
\label{sec:ml}

We shall apply an ML-based scheme to arrive at a selection of MCSs that satisfies some criteria of performance.  As already discussed in the previous section, our scheme involves two ML models, the constructor and evaluator respectively.  The constructor proposes a set of MCSs to the evaluator.  The evaluator assigns a score to this proposal based on design and/or performance criteria that are not known to or shared with the constructor.  After receiving the score from the evaluator in response to its proposed set of MCSs, the constructor now has to update its proposed set of MCSs so as to improve the score to exceed a fixed threshold that was decided in advance between the constructor and evaluator.  

\subsection{Constructor-Evaluator methodology}

\begin{figure*}[htbp]
\centerline{\includegraphics[width=0.9\textwidth]{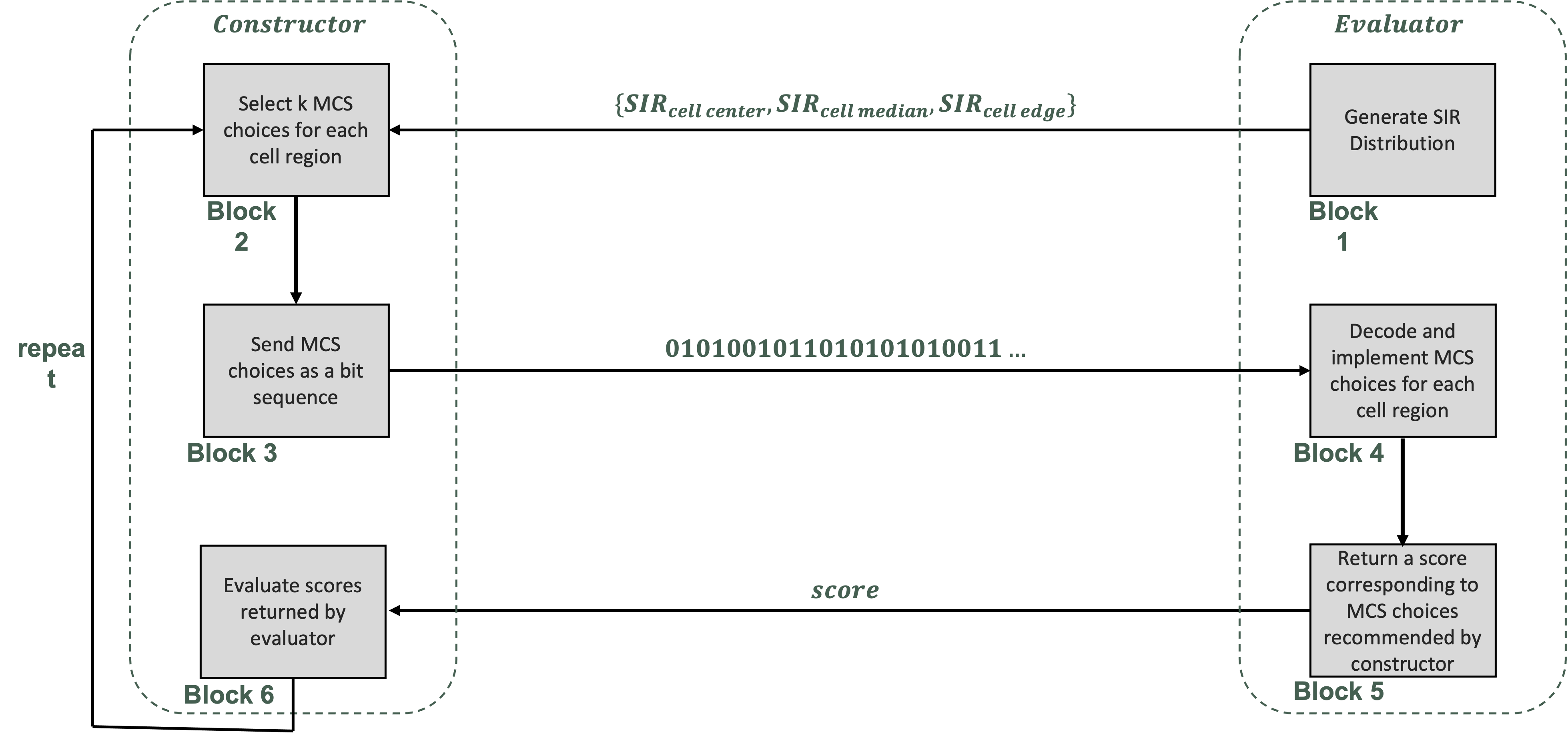}}
\caption{Functional block diagram of the stages of the iterative algorithm with a constructor-evaluator pair whose outcome is a set of $k$ feasible MCSs for each of 3 cell regions: center, median, and edge.}
\label{fig:constr_eval_blk_diag}
\end{figure*}

For an example of this ML methodology, consider a constructor proposal to be a set of MCSs such that there are $k$ possible MCS selections available to an omnidirectional BS transmitting to a served UE in each of three regions of the cell that are not explicitly disclosed to the constructor but implicitly defined by the evaluator (see below): the \emph{cell center}, the \emph{cell edge}, and the region between the cell center and the cell edge, which we shall henceforth call the \emph{cell median}.  Note that the very different mean SIR values for UEs at the cell center and cell edge call for separate MCS sets for each cell region.  The block diagram of the stages of the iterative algorithm that arrives at a ``feasible'' set of MCSs is shown in Fig.~\ref{fig:constr_eval_blk_diag}.  

\subsection{``Feasibility'' of a proposed MCS set}

In our particular use case, we define the ``feasbility'' of an MCS set in terms of SIR-based performance criteria with that MCS set, as detailed below.  The evaluator implicitly defines the cell center, median, and edge to be respectively the 75th, 50th, and 25th percentiles, say, of the SIR distribution of an arbitrarily-located UE.  In practice, this is done by simulating $n$ draws (for some large $n$) from the SIR distribution given by~\eqref{eq:ccdf}, corresponding to $n$ independent identically distributed (iid) UEs located uniformly in the cell, and computing the empirical 25th/50th/75th percentiles of these $n$ draws.

The game between the constructor and evaluator begins with the evaluator sending these 3 percentile values to the constructor (see Block 1 in Fig.~\ref{fig:constr_eval_blk_diag}), signifying the challenge to the constructor to propose (a) a set of $k$ MCSs for a UE whose SIR is each of these percentile values, that (b) achieves a score (given by the evaluator) exceeding some fixed threshold.  We next discuss how the evaluator assigns a score to an MCS set.

\subsection{Score assignment by the evaluator}

For each SIR value set by the evaluator (i.e., the SIR corresponding to a UE located in the cell center, median, or edge respectively), the constructor sends (see Block 3 in Fig.~\ref{fig:constr_eval_blk_diag}) a 15-bit sequence that is just the concatenation of three 5-bit binary strings representing the index (from 0 to 28) of the MCS in the table in Fig.~\ref{fig:mcs_cqi}.  As discussed earlier, knowing the MCS allows the evaluator to use lookup tables like the one in~\cite{b7} to compute the peak throughput, bits per RE, spectral efficiency, and the minimum SNR for viable use of that MCS on that link (see Block 4 in Fig.~\ref{fig:constr_eval_blk_diag}).

Recall that there are $n$ UEs scattered over the cell, and each UE has a (simulated) SIR value.  For each cell region, the evaluator now compares the minimum SNR requirement for each of the $k$ MCSs proposed by the constructor with the (simulated) SIR of each UE in that cell region. For each proposed MCS for that cell region, we count the number of UEs in that cell region whose SIR exceeds the minimum SNR requirement for viable use of that MCS.  The average of this number of UEs across all $k$ proposed MCSs for that cell region is called the \emph{MCS Suitability Score} (MSS) and defines the score assigned by the evaluator (and returned to the constructor) to the set of $k$ proposed MCSs for that cell region (see Block 5 in Fig.~\ref{fig:constr_eval_blk_diag}).  Note that the MSS is always between 0 and 1.

\section{Reinforcement Learning solution}
\label{sec:rl}

We now describe and evaluate an RL solution whereby the constructor can update its set of $k$ MCSs (for each cell region) so as to eventually exceed a fixed threshold for the MSS for that cell region.  Note that we are actually training not one but three RL agents at the constructor, one for each cell region.  

We begin by defining the \emph{state} for each of the RL agents at the constructor.  Although it may be surprising, we will use a quantized version of the MSS for the most recent proposal as the current state.    


Next, we consider the \emph{action space} of each of these RL agents.  If there are $M$ possible MCS choices for a given cell region, then there are $\binom{M}{k}$ actions possible for the RL agent for that cell region, each action being a proposal of a set of $k$ MCSs submitted to the evaluator.  In other words, this RL agent has an action space of size $\binom{M}{k}$.  If we allow the full 29 choices for MCS in each cell region ($M=29$) with $k=3$, say, then $\binom{M}{k} = 3654$, which is impractically large and also unrealistic because, e.g., a UE at the cell edge cannot use the largest MCSs anyway.  Therefore the action space will need to be restricted in some way, as discussed below.  

Lastly, we define the \emph{reward} corresponding to an \emph{action} (defined by proposing a set of $k$ MCSs) as some function of $\mathit{MSS}$ and $\mathit{SE}$, where $\mathit{SE}$ is the average across all UEs in this cell region of the highest spectral efficiency achievable at each UE from amongst the proposed $k$ MCSs.  We may further normalize this $\mathit{SE}$ by the maximum spectral efficiency over all (29) MCSs in order to get an $\mathit{SE}$ between 0 and 1. 

\begin{figure}[htbp]
\centerline{\includegraphics[width=\columnwidth]{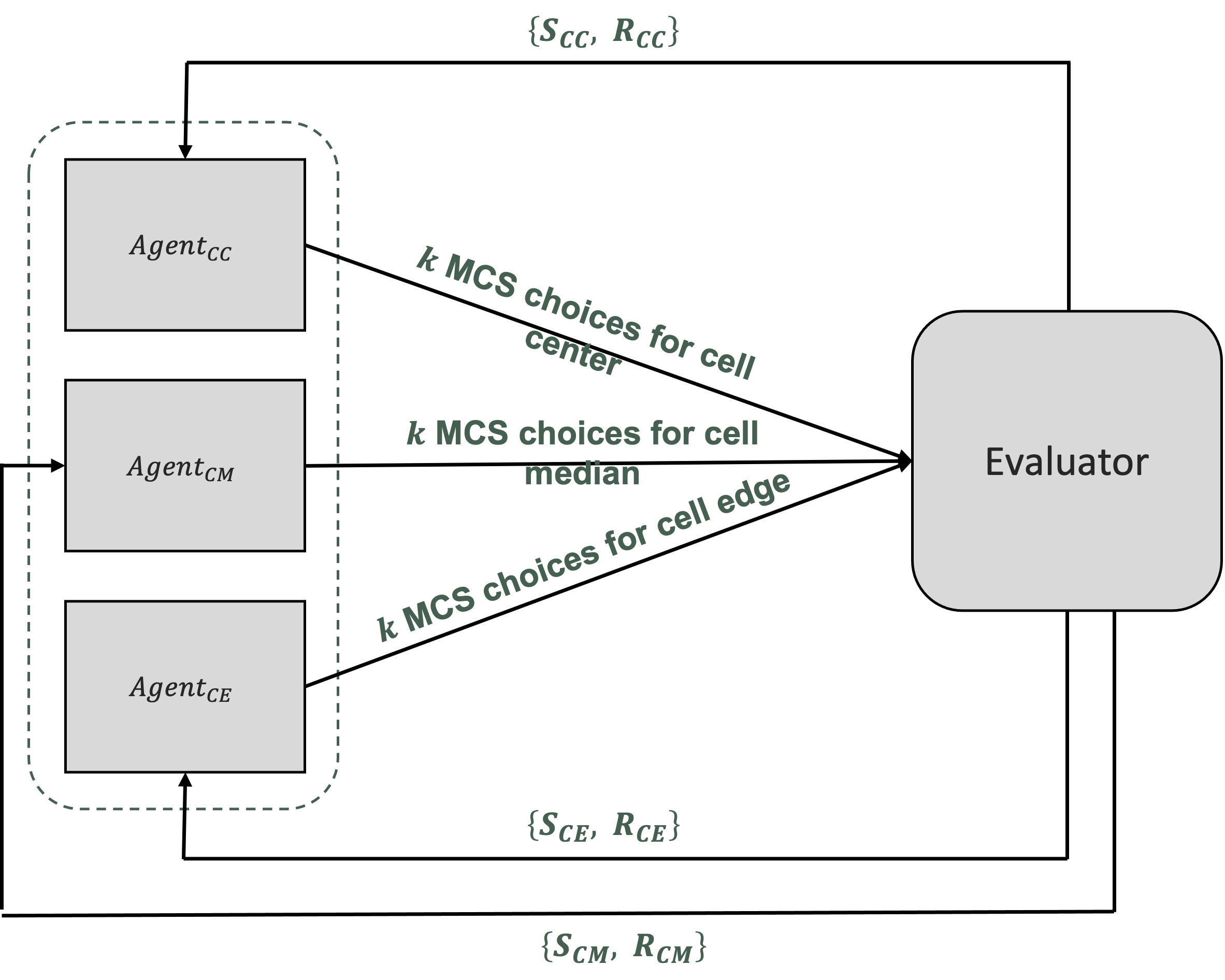}}
\caption{The three RL agents at the constructor, one for each of the cell regions (labeled CC for cell center, CE for cell edge, and CM for cell median), together with the states $S$ and rewards $R$ for each region.}
\label{fig:constr_eval_rl}
\end{figure}

The RL approach is illustrated in Fig.~\ref{fig:constr_eval_rl}.

\subsection{Definition of Action Space}

As discussed above, it is unrealistic and unnecessary to allow each RL agent (for each cell region) to make its selection of $k$ MCSs from the full set of all 29 MCSs.  However, the problem turns out to be unexpectedly subtle.

As a first pass at defining the action space, consider a simple distribution of the 29 MCSs into three sets of roughly equal size, one for each cell region, each set defined by a contiguous selection of MCS indices from the table in Fig.~\ref{fig:mcs_cqi}.  From the point of view of usability of an MCS, we may expect that a UE on the border between two cell regions could probably use both the lowest MCS of one region and the highest MCS of the next, so we allow the three sets above to overlap by one MCS each between adjacent sets.  In other words, the cell edge set of MCSs has $M_{\mathit{CE}}=11$ MCSs corresponding to indices $0,1,\dots,10$, the cell median set of MCSs has $M_{\textit{CM}}=9$ MCSs corresponding to indices $10,11,\dots,18$, and the cell center set of MCSs has $M_{\textit{CC}}=11$ MCSs corresponding to indices $18,19,\dots,28$.  For $k=3$, say, the respective action spaces for the RL agents corresponding to the three cell regions are of sizes $\binom{M_{\mathit{CE}}}{k}=165$, $\binom{M_{\mathit{CM}}}{k}=84$, and $\binom{M_{\mathit{CC}}}{k}=165$ respectively. 

\begin{figure}[htbp]
\centerline{\includegraphics[width=0.8\columnwidth]{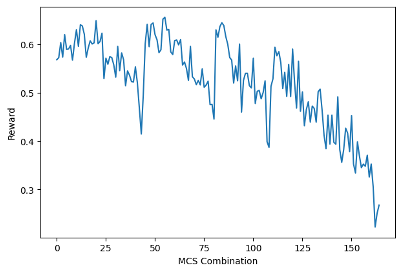}}
\centerline{\includegraphics[width=0.8\columnwidth]{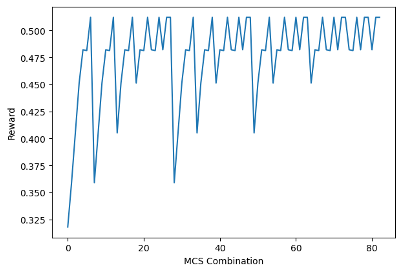}}
\centerline{\includegraphics[width=0.8\columnwidth]{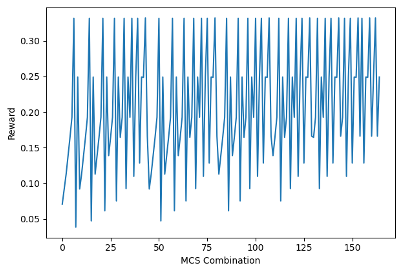}}
\caption{Plots of reward $\mathit{MSS}\,\times\,\mathit{SE}$ vs.~MCS combination in the respective action space for each cell regions (CC at top, CM in middle, CE at bottom) for the naive split of the 29 MCSs into 11, 9, and 11 MCSs in the CC, CM, and CE respectively.  We need a principled selection of $M_{\mathit{CE}}$, $M_{\textit{CM}}$, $M_{\textit{CC}}$, $k$, the way in which we sort and index combinations of $k$ MCSs, and even the functional form of the reward function, in order to get a smoother change in reward when we go from one combination of $k$ MCSs to the next.}
\label{fig:rew_vs_mcs}
\end{figure}

Unfortunately for the above naive construction of MCS sets for the three cell regions, we observe that if we simply index the elements of each of these action spaces (recall that each element of an action space comprises a combination of $k$ MCSs) lexicographically and plot the reward function defined by, say, $\mathit{MSS}\,\times\,\mathit{SE}$ for a cell region versus the index of the member of the action space for that cell region, the plot is not smooth and has multiple optima (see Fig.~\ref{fig:rew_vs_mcs}).  Clearly, RL will fail if trying to find the optimum reward when searching on such a space.
 
\subsection{Reduced Action Space}

We therefore sort and index the MCS combinations in each action space not lexicographically but in increasing order of the total sum of the MCS indices in each combination.  This way, the MCS combinations with ``relatively'' similar modulation and coding rates are placed closer to each other in the enumeration of members of the action space.  We conducted extensive evaluations of different choices of $k$ and splits of the 29 available MCSs among three sets $\mathcal{M}_{\mathit{CC}},\mathcal{M}_{\mathit{CM}},\mathcal{M}_{\mathit{CE}}$ in order to have the smoothest transition of the sum of MCS indices for MCS combinations from $\mathcal{M}_{\mathit{CC}}$ to $\mathcal{M}_{\mathit{CM}}$ and from $\mathcal{M}_{\mathit{CE}}$ to $\mathcal{M}_{\mathit{CE}}$. We also experimented with different mathematical expressions for the reward in order to get a reward vs.~MCS combination plot that is as smooth as possible.  

Finally, we settled on the following: $k=4$ and $M_{\mathit{CC}}=M_{\mathit{CM}}=M_{\mathit{CE}}=12$.
Depending on the typical SIR ranges of each cell region and SINR sensitivity levels from~\cite{b7}, the MCS choices for cell-edge are restricted to MCS indices $\mathcal{M}_{\mathit{CE}}=\{0,1,\dots,11\}$, cell-median are restricted to MCS indices $\mathcal{M}_{\mathit{CM}}=\{6,7,\dots,17\}$, and cell-center are restricted to MCS indices $\mathcal{M}_{\mathit{CC}}=\{17,18,\dots,28\}$.

The number of MCS combinations possible for each cell region is then $\binom{12}{4}=495$.  This is too large a  space to search directly, so we further restrict the possible actions that can be taken by the RL agent at each state to be only one of the following three actions: reuse the present MCS combination, use the preceding MCS combination, or use the succeeding MCS combination, where ``preceding'' and ``succeeding'' are in terms of the indexing of MCS combinations as defined above.

\section{Results and discussion}
\label{sec:results}

In addition to reducing the possible actions of the RL agent for each cell region to just three at each state, we also define a quantized (into one of 20 bins) value of the MSS for that region as the state of the agent, and define the reward as $R = (\mathit{MSS}+\mathit{SE})/2$ for a smoother reward vs.~MCS combination index curve.  Then we run standard Q-learning~\cite[Sec.~6.5]{b6} as described in Algorithm~\ref{alg:ql}.  

\begin{algorithm}[ht]
Initialize: \\
\qquad $\gamma = 0.90$ \\
\qquad $\alpha_{\min}=0.5$, $\alpha_{\mathrm{decay}}=0.995$ \\
\qquad $\epsilon_{\min}=1.0$, $\epsilon_{\mathrm{decay}}=0.995$ \\
\qquad $\alpha \leftarrow \alpha_{\mathrm{init}}=0.7$, $\epsilon \leftarrow \epsilon_{\mathrm{init}}=0.01$ \\
\qquad $Q(s, a)$ for all states $s$ and all actions $a$ arbitrarily\\
Loop for each episode:\\
\qquad Initialize the new state $S$ \\
\qquad Loop for each step of episode:\\
\qquad \qquad Choose $A$ from $\epsilon$-greedy policy from $Q$ \\
\qquad \qquad Take action $A$, get reward $R$, new state $S'$ \\
\qquad \qquad $Q(S,A) \leftarrow $ \\
\qquad \qquad $Q(S,A) + \alpha [R + \gamma \max_a Q(S', a) - Q(S,A)]$ \\
\qquad \qquad $S \leftarrow S'$ \\
\qquad \qquad $\alpha \leftarrow \max(\alpha_{\min}, \alpha \alpha_{\mathrm{decay}})$ \\
\qquad \qquad $\epsilon \leftarrow \max(\epsilon_{\min}, \epsilon \epsilon_{\mathrm{decay}})$ \\
\qquad until $S$ is a terminal state\\
\quad \\
\caption{Q-learning for RL agent for one cell region\label{alg:ql}}
\end{algorithm}

\begin{figure*}[htbp]
\centerline{\includegraphics[width=0.9\textwidth]{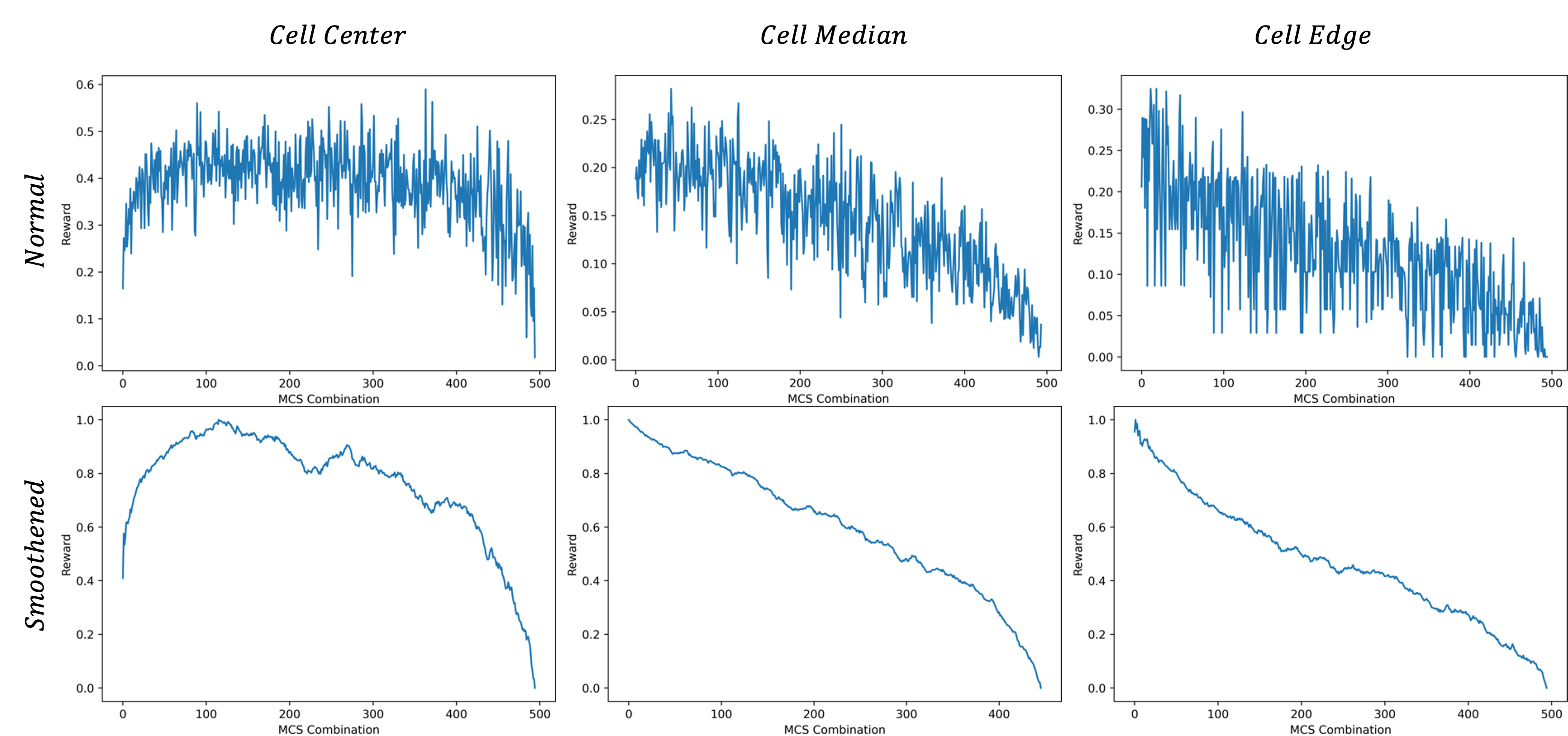}}
\caption{Plots of reward for the three cell regions, unsmoothed $R = (\mathit{MSS}+\mathit{SE})/2$ (top) and $R$ smoothed by a moving average over a rectangular window of size 50 (bottom).  These curves were generated by the brute-force method of exhaustively evaluating all the 495 MCS combinations.}
\label{fig:smoothed_rew_vs_mcs}
\end{figure*}

This reward function is smoothed using a moving average over a rectangular window of width 50.  As is clear from Fig.~\ref{fig:smoothed_rew_vs_mcs}, the plot of the smoothed reward versus MCS combination is fairly smooth for each of the cell regions, thereby making it possible to find a combination of $k=4$ MCSs that maximize accumulated discounted reward. 

\begin{figure}[htbp]
\centerline{\includegraphics[width=0.8\columnwidth]{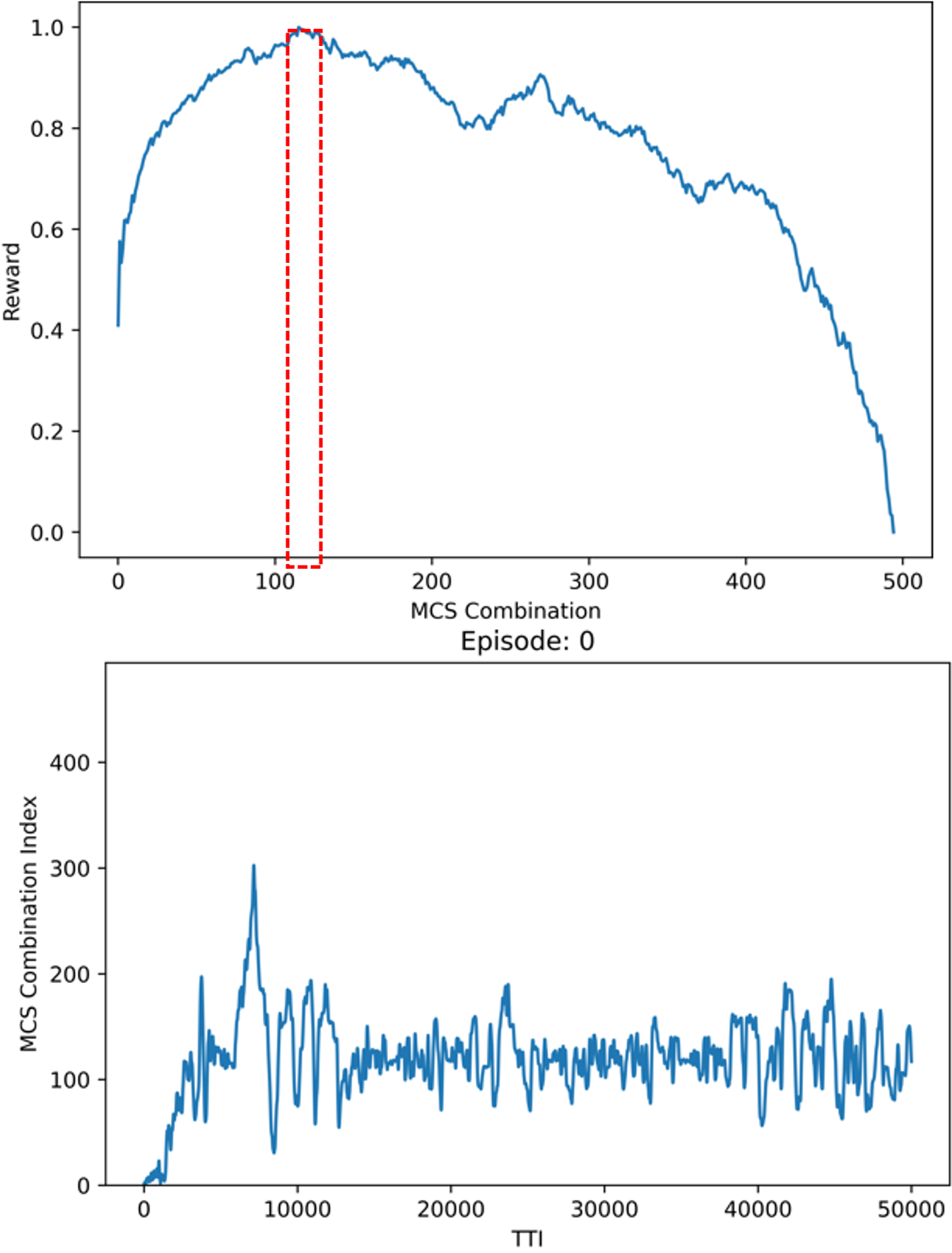}}
\caption{The (bottom) plot of MCS combinations selected by the cell center RL agent after iteration 10000 of the training algorithm shows that the MCS combinations chosen are relatively near in index, and therefore in MSS and SE, and thereby in reward.  This is confirmed from the top plot (same as in Fig.~\ref{fig:smoothed_rew_vs_mcs}), which also shows that the selected MCS combinations all lie within the base of the narrow red rectangle, implying that they are optimal or near-optimal for the (smoothed) reward.}
\label{fig:cc_perf_smoothed}
\end{figure}

Running the Q-learning algorithm shows that the algorithm does ``converge'' to a \emph{terminal} state in the sense that it delivers MCS combination actions that yield values of the smoothed reward function close to its maximum.  In Fig.~\ref{fig:cc_perf_smoothed}, we plot (bottom panel) the MCS combination actions that are selected by the  RL agent for the cell center during a single example episode of RL training between iterations (TTIs) 10000 through 50000, after the training algorithm has converged in the above sense.  Comparing against the plot of the reward function versus MCS combination index from Fig.~\ref{fig:smoothed_rew_vs_mcs} (upper panel) shows that the selected MCS combinations are those for which the (smoothed) reward function is either at or very near its maximum value.  

\begin{figure}[htbp]
\centerline{\includegraphics[width=0.8\columnwidth]{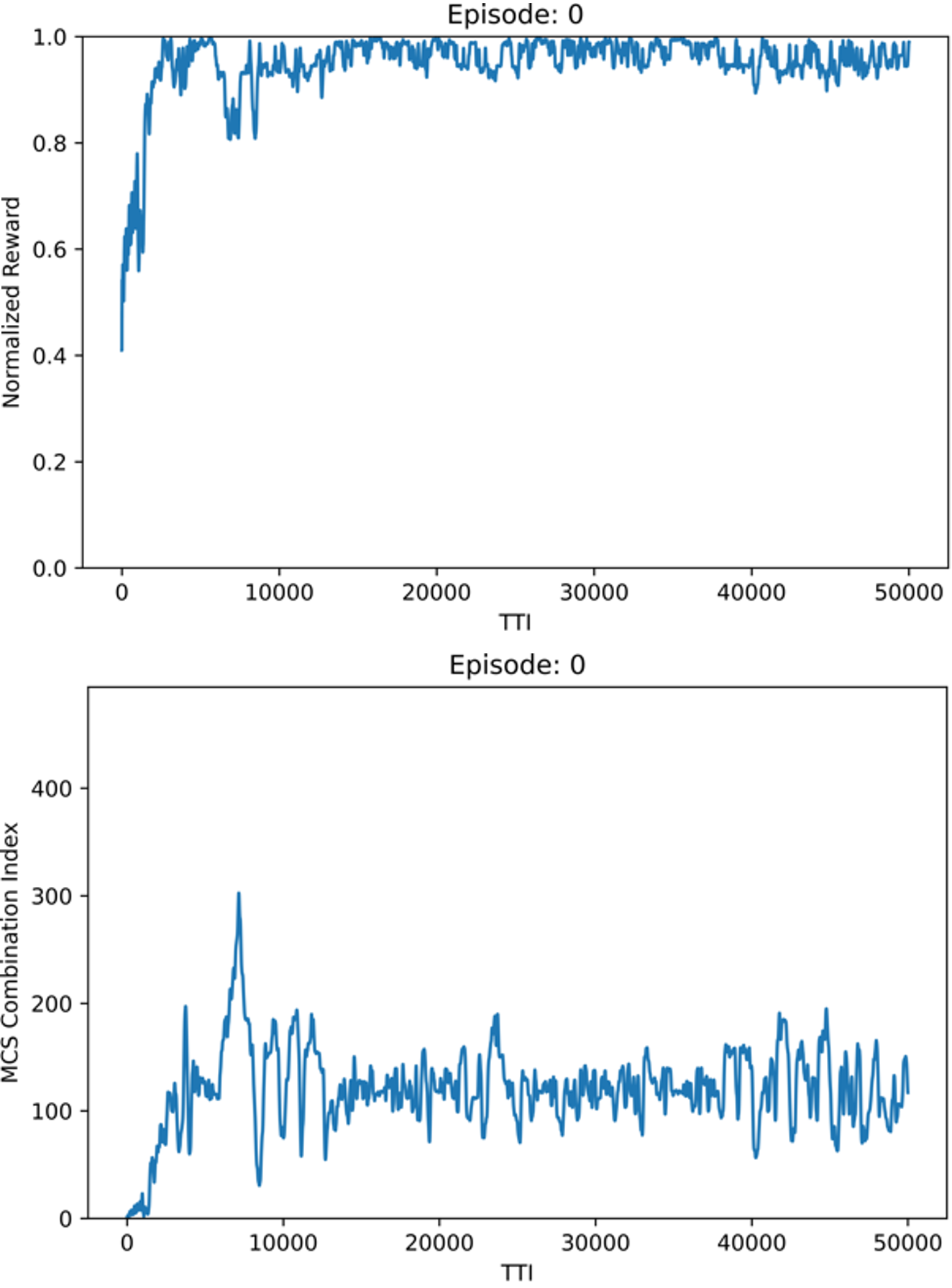}}
\caption{The (bottom) plot of MCS combinations selected by the cell center RL agent after iteration 10000 of the training algorithm is the same as in Fig.~\ref{fig:cc_perf_smoothed}.  The top plot shows that the selected MCS combinations are also optimal or near-optimal for the (unsmoothed) normalized reward.}
\label{fig:cc_perf_norm}
\end{figure}

These observations are further confirmed by plotting the (unsmoothed) normalized reward (see top plot in Fig.~\ref{fig:cc_perf_norm}) corresponding to the same iterations (TTIs) as in the bottom plot in Fig.~\ref{fig:cc_perf_smoothed} for the cell center RL agent.  We observe that the MCS combinations selected by the RL agent after about TTI 10000 are (near-) optimal for the normalized reward as well, since the normalized reward has a maximum value of 1.0.  The same observations as in Fig.~\ref{fig:cc_perf_smoothed} and Fig.~\ref{fig:cc_perf_norm} also hold for the RL agents for the cell median and cell edge regions (plots not shown here for reasons of space).

\section{Conclusions}
\label{sec:concl}

We have proposed a scheme using a pair of constructor-evaluator ML models to automate the selection of MCS sets to satisfy pre-selected design criteria over an LTE-like air interface.  The constructor uses RL in order to train an agent to submit MCS set selections as proposals and iterate based on the scores reported by the evaluator for a proposed submission.  We showed how to design the state space, action space, and select a reward function and demonstrated via simulation that RL does allow the constructor to find an MCS that works.

In general, the selection of the reward function is important, and can be different from the semi-analytical representation we generated here based on lookup table values.  For more complex reward functions, we need a careful optimization of the hyperparameters of the RL algorithm (e.g., Q-learning).

This work shows the potential of ML to automate the tedious, slow and conflict-laden process of standards design in order to enable quick support for new applications.

\end{document}